\documentclass[sigconf]{acmart}

\usepackage{bm}
\usepackage{graphicx}  

\usepackage{amssymb}
\usepackage{subfigure}
\usepackage{amsmath}
\usepackage{gensymb}
\usepackage{booktabs}
\usepackage[normalem]{ulem}

\AtBeginDocument{%
  \providecommand\BibTeX{{%
    \normalfont B\kern-0.5em{\scshape i\kern-0.25em b}\kern-0.8em\TeX}}}

\begin{document}
\fancyhead{}

\title{SSPU-Net: Self-Supervised Point Cloud Upsampling via Differentiable Rendering}

\author{Yifan Zhao$^*$}
\affiliation{%
	\institution{Nanjing University of Science and Technology, China}
	\country{}
}
\email{meidike@outlook.com}

\author{Le Hui$^*$}
\affiliation{%
	\institution{Nanjing University of Science and Technology, China}
	\country{}}
\email{le.hui@njust.edu.cn}

\author{Jin Xie$^{\dagger}$}
\affiliation{%
	\institution{Nanjing University of Science and Technology, China}
	\country{}}
\email{csjxie@njust.edu.cn}

\makeatletter
\def\authornotetext#1{
	\if@ACM@anonymous\else
	\g@addto@macro\@authornotes{
		\stepcounter{footnote}\footnotetext{#1}}
	\fi}
\makeatother
\authornotetext{Equal contribution.}
\authornotetext{Corresponding author.}

\begin{abstract}
Point clouds obtained from 3D sensors are usually sparse. Existing methods mainly focus on upsampling sparse point clouds in a supervised manner by using dense ground truth point clouds. In this paper, we propose a self-supervised point cloud upsampling network (SSPU-Net) to generate dense point clouds without using ground truth. To achieve this, we exploit the consistency between the input sparse point cloud and generated dense point cloud for the shapes and rendered images. Specifically, we first propose a neighbor expansion unit (NEU) to upsample the sparse point clouds, where the local geometric structures of the sparse point clouds are exploited to learn weights for point interpolation. Then, we develop a differentiable point cloud rendering unit (DRU) as an end-to-end module in our network to render the point cloud into multi-view images. Finally, we formulate a shape-consistent loss and an image-consistent loss to train the network so that the shapes of the sparse and dense point clouds are as consistent as possible. Extensive results on the CAD and scanned datasets demonstrate that our method can achieve impressive results in a self-supervised manner. Code is available at https://github.com/fpthink/SSPU-Net.
\end{abstract}



\maketitle

\section{Introduction}

In recent years, as an important 3D data representation, point clouds have received more attention and has been gradually applied to many fields such as autonomous driving~\cite{pham20203d,ding2020lidar,liu2020deep,cao2019adversarial,lu2019l3,wang2019pseudo,zeng2018rt3d}, robotics~\cite{kastner20203d,boroson20193d,zhang2020flowfusion,qin2018detecting,iqbal2020simulation,horvath2017point,liu2016formation} and virtual reality~\cite{blanc2020genuage,stets2017visualization,bonatto2016explorations,xu2019automatic}. Generally, the raw point clouds obtained by 3D sensors are usually sparse. Sparse point clouds cannot characterize local geometric structures of 3D objects well, which may lead to the poor performance of the downstream point cloud processing tasks such as 3D object segmentation, detection and classification. 
Therefore, it is necessary to upsample sparse point clouds to generate dense and complete point clouds that can facilitate the subsequent point cloud processing tasks.

In the past decades, some optimization based upsampling methods~\cite{huang2013edge,wu2015deep} were proposed to upsample point clouds, which usually rely on geometric priors, such as normal estimation and smooth surfaces. Nonetheless, the complex geometric structures of 3D objects usually limit performance of these methods. Recently, a lot of deep learning based supervised methods~\cite{pu-net,ec-net,pppu,pu-gan,qian2020pugeo,ye2021meta} were proposed to tackle the point cloud upsampling task. As a pioneering work, point cloud upsampling network (PU-Net)~\cite{pu-net} proposed a network paradigm for point cloud upsampling, including three stages of feature extraction, feature augmentation, and coordinate restoration. To improve the quality of upsampled point clouds, multi-step progressive upsampling (MPU) network~\cite{pppu} used a cascaded patch-based upsampling network to progressively upsample point clouds on different levels of point clouds in an end-to-end manner. Lately, based on generative adversarial network (GAN), point cloud upsampling adversarial network (PU-GAN)~\cite{pu-gan} learned the distribution of points from the latent space and upsampled points on the surface patches of the object. These methods can achieve impressive point cloud upsampling results in a supervised manner by using dense ground truth point clouds. Nonetheless, it is difficult to acquire dense ground truth point clouds in the real scenarios. Therefore, it is desirable to develop self-supervised/unsupervised point cloud upsampling methods without using ground truth.

\begin{figure*}[htbp]
	\begin{center}
		\includegraphics[width=1.0\linewidth]{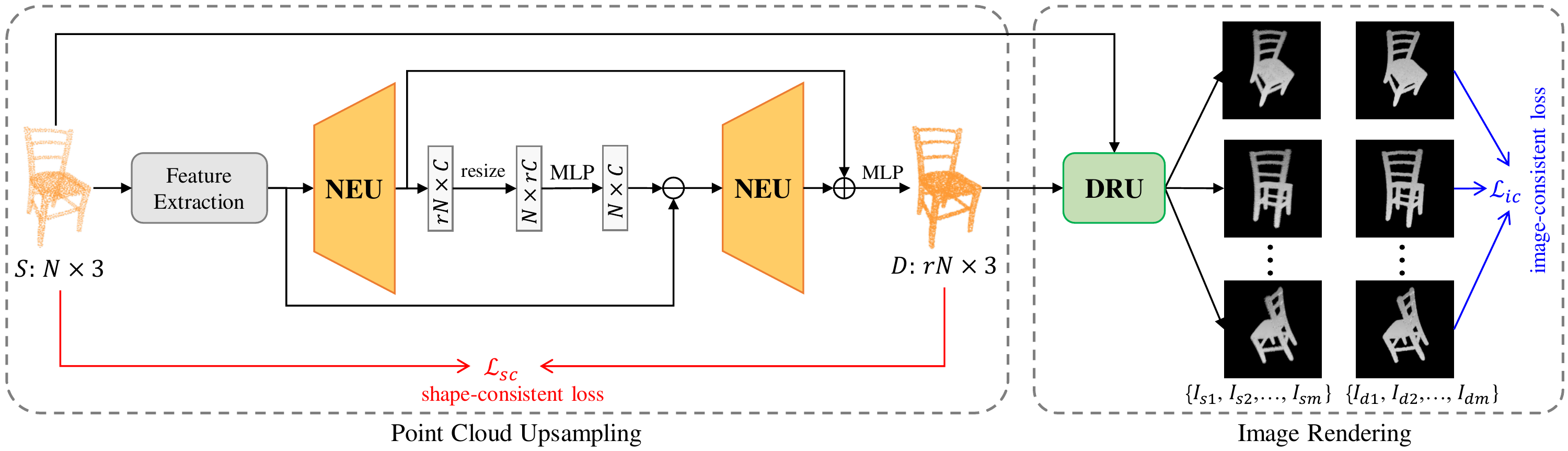}
	\end{center}
	\caption{Overview of the self-supervised point cloud upsampling network (SSPU-Net). The network inputs the sparse point cloud $\bm{S}$ containing $N$ points, and outputs the dense point cloud $\bm{D}$ after being expanded by the neighbor expansion unit (NEU). Note that $r$ is the upsampling rate. Consequently, we adopt a differentiable rendering unit (DRU) to generate multi-view rendering images from the obtained dense point cloud. To train our network in a self-supervised manner, we propose a shape-consistent loss $\mathcal{L}_{sc}$ and an image-consistent loss $\mathcal{L}_{ic}$ to encourage the shapes of the sparse and dense point clouds to be consistent.}
	\label{fig:network}
\end{figure*}

In this paper, we propose a self-supervised point cloud upsampling network (SSPU-Net), which can be trained in an end-to-end manner and does not require ground truth. To enable self-supervised learning, we encourage the input sparse point cloud and the generated dense point cloud to have similar geometric structures by imposing consistency constraints on the 3D shapes and rendered images of the sparse and dense point clouds.
Specifically, we first propose a neighbor expansion unit (NEU) to upsample the sparse point cloud to the dense point cloud, where we exploit the local geometric structures of each point to adaptively learn weights to interpolate the features of new points. Besides, we also adopt the 2D grid mechanism in~\cite{pu-gan} to generate a unique 2D vector for each point feature in the NEU. After stacking two NEUs, we use a multi-layer preceptron (MLP) to generate the 3D coordinates of the dense point cloud. Then, we develop a differentiable rendering unit (DRU) to render the input sparse point cloud and the generated dense point cloud into multi-view images. In order to make the sparse and dense point clouds have similar geometric structures, we formulate a shape-consistent loss on point clouds and an image-consistent loss on rendered images to train the point cloud upsampling network. Experimental results on the CAD and scanned datas demonstrate that our SSPU-Net can achieve impressive results in a self-supervised manner. In comparison to the supervised point cloud upsampling methods, our upsampling results are even better than PU-Net~\cite{pu-net}. In addition, we also conduct experiments on the downstream point cloud classification and point cloud based place recognition tasks. The experimental results further show that the upsampled point clouds by our self-supervised method can improve the performance of these tasks.

The main contributions of the paper are as follows:
\begin{enumerate}
	\item We propose a neighbor extended unit (NEU) to upsample the sparse point clouds by adaptively learning weights from local geometric structures of the point clouds. We also develop a differentiable rendering unit (DRU) to render the point clouds into multi-view images, which can be nested in the network for end-to-end training.
	
	\item We formulate a shape-consistent loss and an image-consistent loss to make the sparse and dense point clouds consistent so that the point cloud upsampling network can be trained in a self-supervised manner.
	
	\item Our method can achieve impressive results in a self-supervised manner, and outperforms some supervised point cloud upsampling methods.
\end{enumerate}

\section{Related Work}

\begin{figure*}
	\begin{center}
		\includegraphics[width=1.0\linewidth]{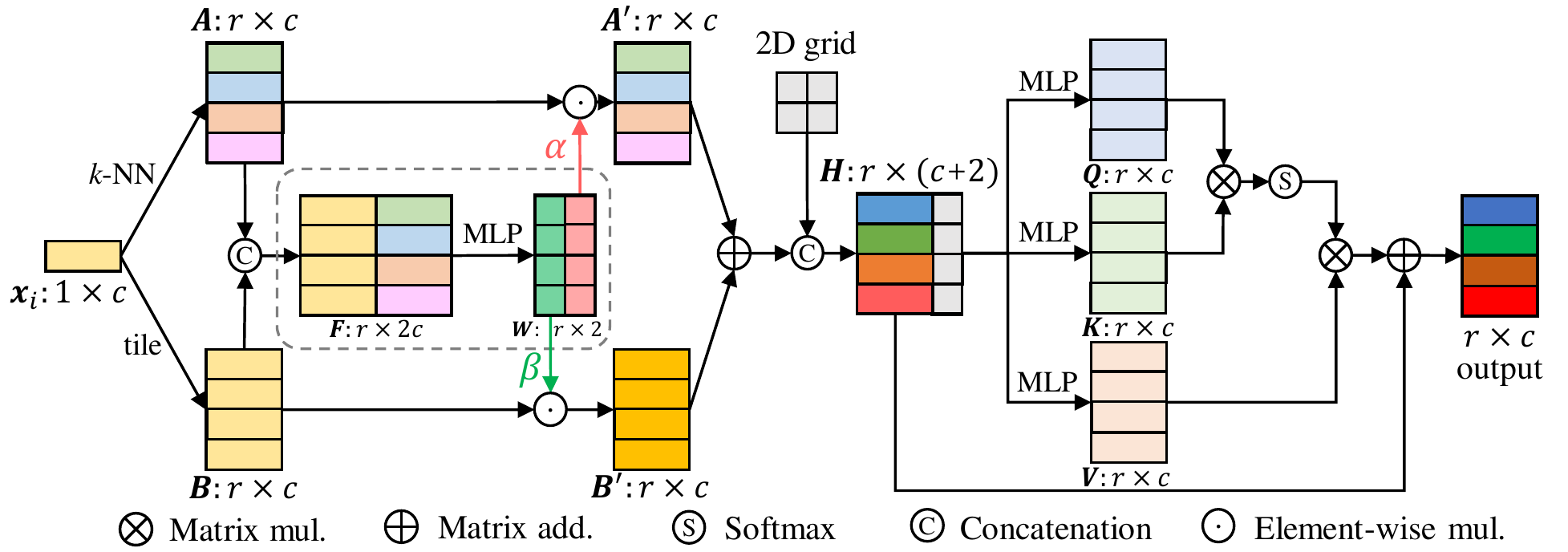}
	\end{center}
	\caption{Illustration of the neighbor expansion unit (NEU). Given the point feature $\bm{x}_i\in\mathbb{R}^c$, we first use the $k$-nearest neighbor to construct the local feature map $\bm{A}\in\mathbb{R}^{r\times c}$, and repeat the point feature $\bm{x}_i$ for $r$ times to generate the feature map $\bm{B}\in\mathbb{R}^{r\times c}$. Then, we adaptively learn the weights $\bm{\alpha}\in\mathbb{R}^{r}$ and $\bm{\beta}\in\mathbb{R}^{r}$ from the concatenated feature map $\bm{F}\in\mathbb{R}^{r\times 2c}$. Consequently, we use $\bm{\alpha}$ and $\bm{\beta}$ to weight the feature maps $\bm{A}$ and $\bm{B}$ to obtain a new feature map. Finally, we use the 2D grid mechanism and the self-attention mechanism to generate the final feature map.}
	\label{fig:Expansion Unit}
\end{figure*}

\textbf{Optimization based point cloud upsampling.} Generally, optimization based methods use geometric priors, such as normal estimation and smooth surfaces, to upsample the sparse point cloud to the dense point cloud. Alexa~\emph{et al.}~\cite{alexa2003computing} interpolated new points at the vertex of the Voronoi diagram of the local tangent plane of the surfaces to implement point cloud upsampling. In order to get rid of local plane fitting, Lipman~\emph{et al.}~\cite{lipman2007parameterization} proposed a local optimization mapping operator (LOP) to approximate the surface, in which we can resample the point cloud to obtain the dense point cloud. Likewise, Huang~\emph{et al.}~\cite{huang2013edge} developed an edge-aware point set resampling method, called EAR, where the LOP is used to push points away from the edges. Recently, Wu~\emph{et al.} proposed a consolidation method~\cite{wu2015deep} based on deep point representation to fill in the points in the missing areas. To implement point cloud upsampling, it uses the meso-skeleton of the 3D object to resample the points on the surface. Although these methods can achieve good upsampling results, they heavily rely on handcrafted geometric priors of the point cloud. For complex geometric structures of 3D objects, the performance of these methods is limited by the poor geometric priors.

\textbf{Learning based point cloud upsampling.} With the success of deep learning on point clouds, numerous efforts have been made for point cloud processing, including classification~\cite{qi2017pointnet,qi2017pointnet++,dgcnn,mao2019interpolated}, semantic segmentation~\cite{shen2018mining,zhao2019pointweb,lin2020convolution,cheng2020cascaded,cheng2021sspc}, completion~\cite{yuan2018pcn}, generation~\cite{yang2019pointflow,shu20193d,hui2020progressive}. 
Based on PointNet++~\cite{qi2017pointnet++}, Yu~\emph{et al.}~\cite{pu-net} proposed a point cloud upsampling network (PU-Net), which is the first deep learning based point cloud upsampling method. It first learns multi-level features of each point and then executes feature expansion to expand point features. Finally, the expanded point feature is used to generate the 3D coordinates of the dense point cloud. Based on PU-Net, edge-aware point set consolidation network (EC-Net)~\cite{ec-net} learns edge-aware features to better capture the local geometric structures of the point cloud. Although EC-Net can achieve higher performance than PU-Net, it additionally requires the triangular surface annotation of point clouds. Lately, Wang~\emph{et al.}~\cite{pppu} proposed a patch-based progressive upsampling network (MPU), which can capture detailed

structures at different levels of point clouds through a cascaded upsampling network. Li~\emph{et al.}~\cite{pu-gan} proposed point cloud upsampling adversarial network (PU-GAN), which uses the generative adversarial network to improve the quality of the upsampled point cloud with the adversarial training strategy. Although the above methods have achieved impressive results, they all require ground truth for supervised training.

\textbf{Differentiable rendering.} In order to obtain 2D representation of the corresponding 3D objects, many existing methods utilize rendering techniques, such as 3D surface reconstruction~\cite{tewari2017mofa,genova2018unsupervised,tran2018nonlinear,tewari2018self}, material inference~\cite{liu2017material,deschaintre2018single}, and 3D reconstruction tasks~\cite{zienkiewicz2016real,kundu20183d,henderson2018learning}. Loper~\emph{et al.}~\cite{loper2014opendr} proposed an approximate differentiable renderer (DR), which explicitly learns the relationship between changes in model parameters and image observations. Latter, Kato~\emph{et al.}~\cite{kato2018neural} developed a hand-crafted function that can approximate the backward gradient of rasterization to achieve differentiable rendering. Likewise, Li~\emph{et al.} proposed a method~\cite{li2018differentiable} to realize the differentiability of the secondary rendering effect, which uses a differentiable Monte Carlo ray tracer. Lately, Liu~\emph{et al.}~\cite{liu2019soft} proposed a differentiable mesh rendering method called Soft Rasterizer, which can directly render colorized meshes using differentiable functions. In this paper, we use the differentiable rendering method to render point clouds into multi-view images.

\section{Method}
Assuming that a sparse point cloud $\bm{S}=\left\{\bm{s}_i\right\}_{i=1}^N$ has $N$ points, where $\bm{s}_i$ is the 3D coordinates, the goal of our self-supervised method is to generate a dense point cloud $\bm{D}=\{\bm{d}_i\}_{i=1}^{rN}$ ($r$ is the upsampling rate), where $\bm{d}_i$ is the upsampled 3D coordinates and $rN$ is the number of the dense point cloud. As shown in Figure ~\ref{fig:network}, we illustrate the framework of our self-supervised point cloud upsampling network (SSPU-Net). The framework mainly consists of two modules: a neighbor expansion unit (NEU) and a differentiable rendering unit (DRU). In our SSPU-Net framework, we use the sparse point cloud as the input for feature extraction. Following MPU~\cite{pppu}, we use the dense feature extraction unit to extract the local features of the sparse point clouds. Then, we feed the obtained point features into two stacked NEUs to adaptively learn weights from the local geometric structures of point clouds for point interpolation. To obtain the upsampled 3D point cloud, we use a multi-layer perceptron (MLP) to generate the 3D coordinates of the dense point cloud. After obtaining the upsampled point cloud, in the rendering stage, we use the DRU to render the input sparse point cloud and the generated dense point cloud into multi-view images. To train our self-supervised method, we formulate a shape-consistent loss and an image-consistent loss to encourage the shapes of the sparse and dense point clouds to be consistent. Since the DRU is differentiable, the network can be trained in an end-to-end manner.

\subsection{Neighbor Expansion Unit}
Given a learned feature map $\bm{X}=[\bm{x}_1, \bm{x}_2,\cdots, \bm{x}_N]$ of the sparse point cloud $S$ and the expansion factor $r$, the neighbor expansion unit (NEU) aims to learn a dense feature map $\bm{Y}=[\bm{y}_1, \bm{y}_2,\cdots, \bm{y}_{rN}]$. Existing methods such as PU-Net~\cite{pu-net}, MPU~\cite{pppu} and PU-GAN~\cite{pu-gan} mainly execute feature expansion by duplicating point features without considering the local geometric structures of the point cloud. Since ground truth point clouds are used to train the network, they can achieve good upsampling results. However, in the self-supervised upsampling task, the feature expansion operations in these methods cannot generate high-quality point features due to the unavailable ground truth point cloud. To this end, we propose the NEU to adaptively learn weights from the local geometric structures of the point cloud for point interpolation.



As illustrated in Figure~\ref{fig:Expansion Unit}, we depict the detailed structure of the NEU. Given the $i$-th point feature $\bm{x}_i\in\mathbb{R}^{c}$, where $c$ is the feature dimension, we first construct the local neighborhood $\mathcal{N}_i$ with the $r$-nearest points in the spatial space to construct the feature map $\bm{A}\in\mathbb{R}^{r\times c}$, where the integer $r$ is the upsampling rate. At the same time, we repeat the feature vector $\bm{x}_i$ for $r$ times to generate the feature map $\bm{B}\in\mathbb{R}^{r\times c}$. By concatenating feature maps $\bm{A}$ and $\bm{B}$, we can obtain the new feature map $\bm{F}=[\bm{A};\bm{B}]$, where $[\cdot;\cdot]$ represents the channel-wise concatenation and $\bm{F}\in\mathbb{R}^{r\times 2c}$. In this way, we construct the local neighborhood per point. After that, we utilize the multi-layer perceptron (MLP) on the feature map $\bm{F}$ to adaptively learn the weight $\bm{W}\in\mathbb{R}^{r\times 2}$ for point interpolation. As shown in Figure~\ref{fig:Expansion Unit}, we split the weight $\bm{W}\in\mathbb{R}^{r\times 2}$ into two parts of the weight $\bm{\alpha}\in\mathbb{R}^{r}$ and the weight $\bm{\beta}\in\mathbb{R}^{r}$. $\bm{\alpha}$ and $\bm{\beta}$ represent the weights of the adjacent points and center points, respectively. Then, the point interpolation is formulated as:
\begin{equation}
\bm{h}_j = \gamma(\alpha_j)\cdot \bm{x}_i + \gamma(\beta_j)\cdot \bm{x}_j
\end{equation}
where $\bm{h}_j\in\mathbb{R}^{c}$ is the interpolated point feature by weighting the point feature $\bm{x}_i$ and $j$-th point feature $\bm{x}_j$, where $j$ enumerates the index of points in $\mathcal{N}_i$. Besides, $\gamma(\cdot)$ is the \emph{sigmoid} activation function. As a result, for the point feature $\bm{x}_i$, we can generate $r$ new point features. Thus, we upsample the point cloud by $r$ times. Following~\cite{pu-gan}, we also use a 2D grid mechanism and a self-attention mechanism for better feature fusion.

In the experiment, we stack two NEUs to upsample the point cloud. After the first NEU, we first resize the feature size from $rN\times C$ to $N\times rC$ and then apply MLP to generate a new feature map with the size $N\times C$. Consequently, we subtract the new feature map with the original feature map $\bm{X}$ and then feed it into the second NEU. Finally, we use the MLP network to generate the 3D coordinates of the upsampled point cloud. 

\begin{figure}
	\begin{center}
		\includegraphics[width=1\linewidth]{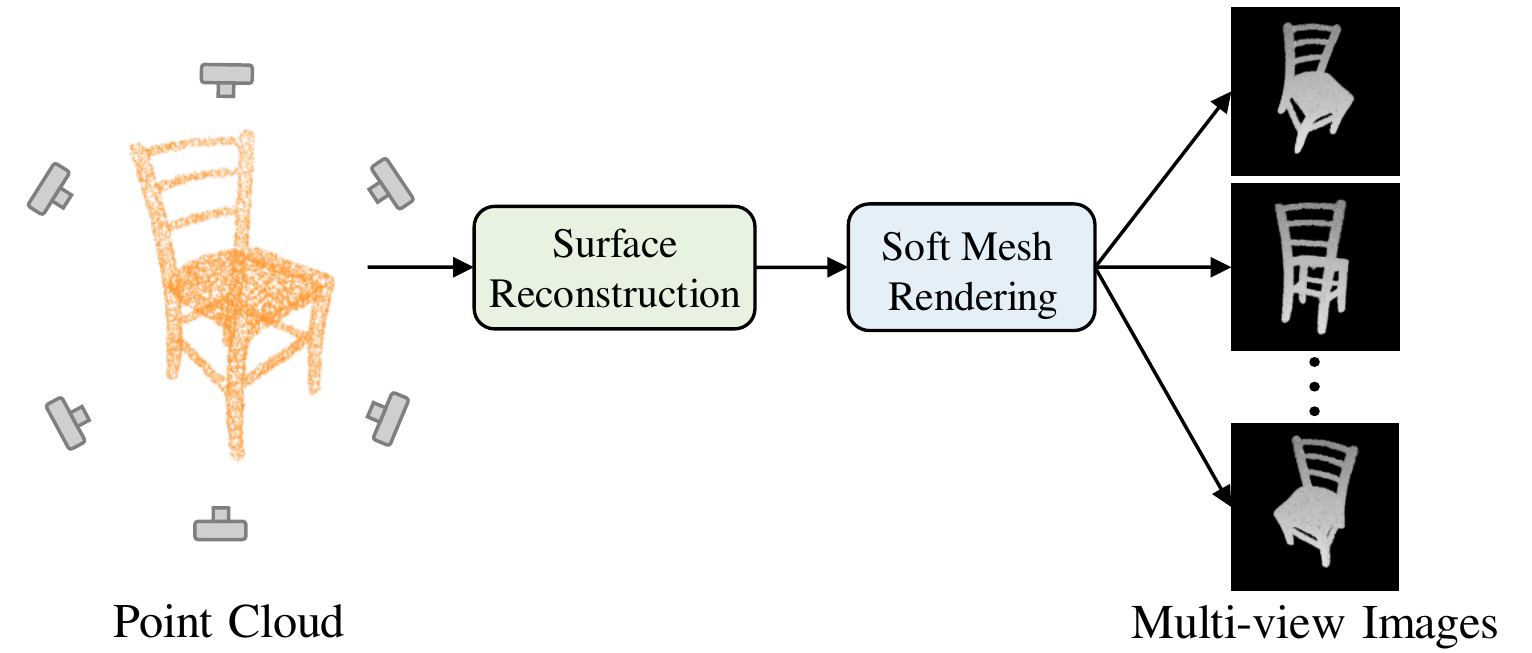}
	\end{center}
	\caption{Illustration of the differentiable rendering unit (DRU).}
	\label{fig:Project Render}
\end{figure}

\subsection{Differentiable Rendering Unit}
After generating the upsampled point cloud, we develop a differentiable rendering unit (DRU) to render the sparse and dense point clouds into multi-view images for computing the subsequent image-consistent loss. As shown in Figure~\ref{fig:Project Render}, we illustrate the differentiable rendering unit. Specifically, we first place the virtual cameras at different viewpoints around the input point cloud to reconstruct the projection surfaces of the point cloud. We then utilize the soft mesh rendering method~\cite{liu2019soft} to render multi-view images from the projection surfaces.

\begin{figure}
	\begin{center}
		\includegraphics[width=0.92\linewidth]{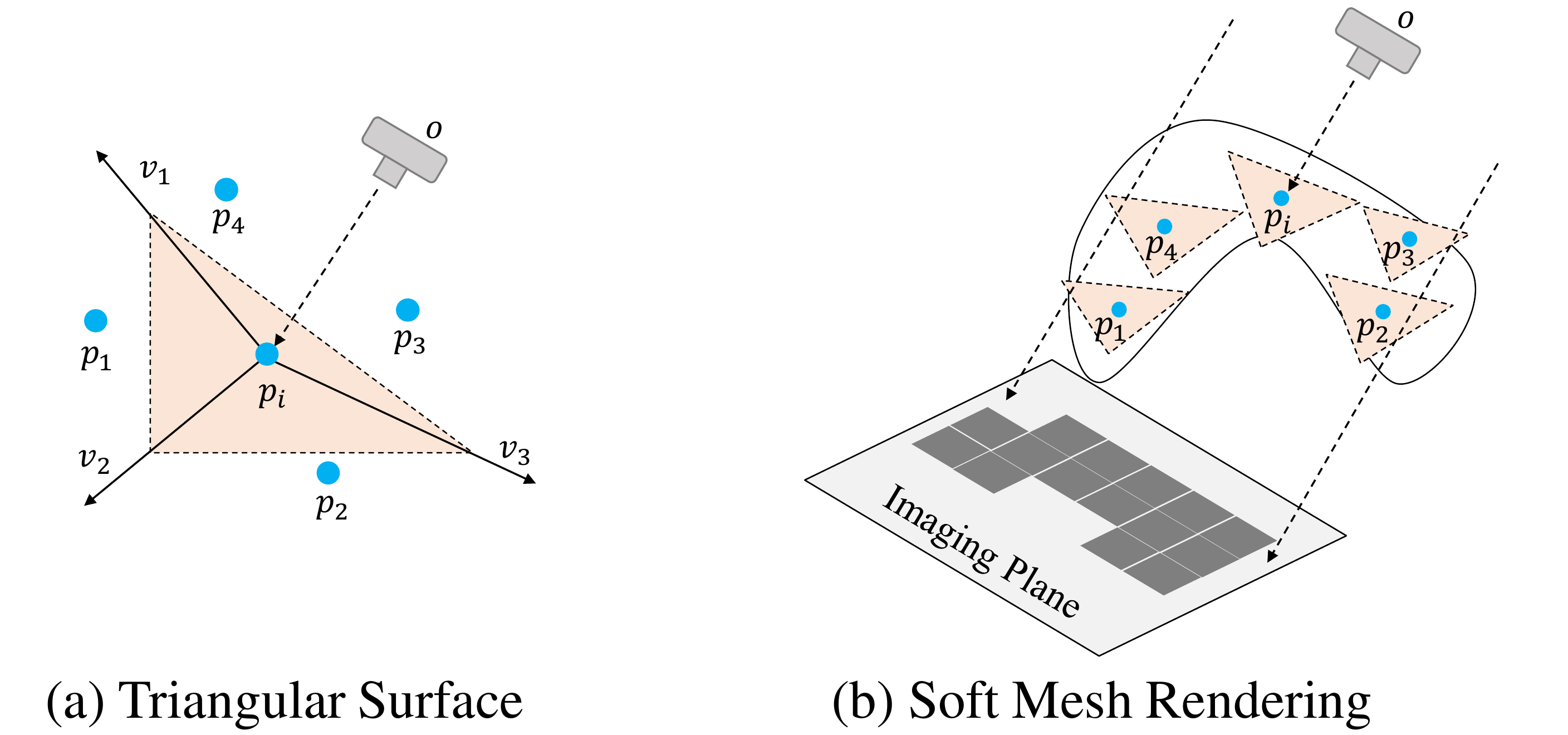}
	\end{center}
	\caption{Illustration of the local projection surfaces of the point cloud.}
	\label{fig:Surface render}
\end{figure}

To obtain the project surfaces of the point cloud, we present an approximate approach to construct a local tangent triangular plane for each point. As shown in Figure~\ref{fig:Surface render} (a), given the point $\bm{p}_i\in\mathbb{R}^3$ and the camera center $\bm{o}\in\mathbb{R}^3$, we can obtain the projection vector $\bm{x}=\bm{p}_i-\bm{o}$ by subtracting the point $\bm{p}_i$ with the camera center $\bm{o}$. Then, we compute the vectors $\bm{v}_1$, $\bm{v}_2$, and $\bm{v}_3$ for constructing the local tangent triangular plane. 
For a unit vector $\bm{s} = [1, 0, 0]$, we can compute the vector $\bm{v}_1$ by:
\begin{equation}
\bm{v}_1 = \bm{x}\times\bm{s}
\label{v1}
\end{equation}
In order to compute the vectors $\bm{v}_2$ and $\bm{v}_3$, we first calculate an auxiliary vector $\hat{\bm{v}}$ perpendicular to vectors $\bm{s}$ and $\bm{v_1}$ by:
\begin{equation}
\hat{\bm{v}} = \bm{s}\times\bm{v}_1
\end{equation}
After that, we can calculate the vector $\bm{v}_2$ by solving the following equations: 
\begin{equation}
\begin{cases}
\bm{v}_2 \times \bm{v}_1 \cdot \bm{\hat{v}} = 0\\
\bm{\hat{v}} \cdot \bm{v}_2 =  |\bm{v_2}||\bm{\hat{v}}|\cos30\degree\\
|\bm{v}_2 \times \bm{\hat{v}}| = |\bm{v}_2||\bm{\hat{v}}|\sin30\degree\\
\bm{v}_1 \cdot \bm{v}_2 = |\bm{v_1}||\bm{v}_2|\cos120\degree
\end{cases}
\label{eqn:v23}
\end{equation}
Note that we can also compute the vector $\bm{v}_3$ by replacing the vector $\bm{v}_2$ in Equation~(\ref{eqn:v23}) with the vector $\bm{v}_3$. Finally, the angles between the vectors $\bm{v}_1$, $\bm{v}_2$ and $\bm{v}_3$ are 120 degrees. Besides, we normalize the vectors $\bm{v}_1$, $\bm{v}_2$, and $\bm{v}_3$ to unit vectors. As a result, we construct the tangent triangular plane for each point. 
Once we construct the local tangent triangular plane, we feed the constructed approximate surface to the differential mesh rendering method~\cite{liu2019soft} to render the final images. Besides, Figure~\ref{fig:Surface render} (b) shows an example of rendering the approximate surface into an image.

\subsection{Loss Functions}

\textbf{Shape-consistent loss.} Generally, the Chamfer Distance (CD) or Earth Mover's Distance (EMD) are used for the point cloud upsampling task. Due to the dense ground truth point clouds, they can work well in a supervised manner. Since the EMD requires that the number of points on the two input point clouds should be consistent, it cannot be applied to our self-supervised method. Compared to EMD, the CD does not require the input point clouds to have the same number of points. Nonetheless, due to the uneven distribution between the input sparse point cloud and the generated dense point cloud, the CD does not work well. Therefore, we propose a shape-consistent loss to make the shape of the dense point cloud consistent to the shape of the sparse point cloud. Specifically, we regard the sparse point cloud $\bm{S}$ as the result of downsampling the dense point cloud $\bm{D}$, so the downsampling result $\hat{\bm{D}}$ shares the common local geometric structures with $\bm{D}$. Given a downsampling function $g(\cdot)$, we can obtain the downsampling point cloud $\hat{\bm{D}}$ from the dense point cloud $\bm{D}$.
By minimizing the EMD between the input sparse point cloud $\bm{S}$ and the downsampling point cloud $\hat{\bm{D}}$, we enforce the dense point cloud to have the consistent shape with the sparse point cloud. 
The proposed shape-consistent loss is formulated as:
\begin{equation}
\mathcal{L}_{sc} = \min_{\psi:\hat{\bm{D}}\to{\bm{S}}}\sum_{\bm{d}_i\in \hat{\bm{D}}}\lVert \bm{d}_i-\psi(\bm{d}_i)\rVert^2_F
\end{equation}
where $\hat{\bm{D}} = g(\bm{D})$. Note that there are many different choices for the downsampling function $g(\cdot)$, and we use the farthest point sampling (FPS) for simplicity in the experiment.

\textbf{Image-consistent loss.} After rendering the sparse and dense point clouds into multi-view images, we formulate an image-consistent loss to make the shapes of the input sparse point cloud and the generated dense point cloud consistent. Given the sparse point cloud $\bm{S}$, we can obtain the rendered images $\{\bm{I}_{s1}, \bm{I}_{s2}, \ldots, \bm{I}_{sm}\}$ at different camera poses, where $m$ is the number of the cameras. Similarly, we can obtain the rendered images $\{\bm{I}_{d1}, \bm{I}_{d2}, \ldots, \bm{I}_{dm}\}$ of the dense point cloud $\bm{D}$. As the dense point cloud has more points than the sparse point cloud, the rendered images of the dense point cloud have more triangular surfaces. Nonetheless, according to experimental results, this gap has few effects on upsampling.
We encourage the shapes of the sparse and dense point clouds to be as consistent as possible, and formulate an image-consistent loss, which is written as:
\begin{equation}
\mathcal{L}_{ic} =\frac{1}{m} \sum_{j=1}^{m}{||\bm{I}_{sj}-\bm{I}_{dj}||}^2_F
\end{equation}
where $\bm{I}_{sj}$ and $\bm{I}_{dj}$ are the rendered images of the sparse and dense point clouds, respectively.



The joint loss function to train the network is defined as:
\begin{equation}
\mathcal{L} = \lambda_{sc} \mathcal{L}_{sc} + \lambda_{ic} \mathcal{L}_{ic} + \lambda_{hd} \mathcal{L}_{hd} + \lambda_{un} \mathcal{L}_{un}
\end{equation}
where $\mathcal{L}_{sc}$ and $\mathcal{L}_{ic}$ are the proposed shape-consistent loss and the image-consistent loss, respectively. Note that $\mathcal{L}_{hd}$ is the Hausdorff distance loss that is used to constrain the noise points and outliers, and $\mathcal{L}_{un}$ is the uniform loss in PU-GAN~\cite{pu-gan} to make the generated point cloud as uniform as possible. For more details of the uniform loss, please refer to~\cite{pu-gan}. $\lambda_{sc}$, $\lambda_{ic}$, $\lambda_{hd}$ and $\lambda_{un}$ are hyperparameters.

\section{Experiment}
\subsection{Dataset and Implementation Details}
We collect datasets released by PU-Net~\cite{pu-net}, MPU~\cite{pppu}, PU-GAN~\cite{pu-gan}, and part of human body models in FAUST~\cite{bogo2014faust} to construct a large dataset, called SSPU-DataSet, which includes 227 3D models of different categories. We randomly select 191 models as the training set, and the remaining models as the test set. We use the same strategy as in MPU~\cite{pppu} to extract 195 patches per 3D object, so a total of 37,245 patches are obtained. Besides, we set 256 points per patch. For a fair comparison, we also conduct experiments on the dataset used in PU-GAN~\cite{pu-gan}. During testing, we use the Chamfer distance (CD), Hausdorff distance (HD), and point-to-surface (P2F) distance to evaluate the performance of different methods.

During training, we use data augmentation to prevent the network from overfitting. We use Adam~\cite{kingma2014adam} as the optimizer and set the learning rate to 0.001. In the experiment, we train the network for 30 epochs with a batch size of 28. 
In NEU, we set $k$=4 to construct the local neighborhoods of point clouds. The hyperparameters $\lambda_{sc}$, $\lambda_{ic}$, $\lambda_{hd}$, and $\lambda_{un}$ are set to 100, 30, 10, and 25, respectively. The whole experiments are conducted on a single 2080Ti GPU.

\subsection{Results}
\textbf{Quantitative results.} As shown in Table~\ref{tab:comparisons2048}, we report the upsampling results of different methods on our SSPU-Dataset. Note that we have listed the results of two cases of inputting 2048 points and 4096 points, respectively. It can be seen that our SSPU-Net has achieved comparable results to those supervised methods such as MPU and PU-GAN. It is worth noting that our SSPU-Net (self-supervised) outperforms PU-Net (supervised) with all metrics. Although our method is self-supervised, our method considers the local geometric structures of the point cloud, while PU-Net does not.
\begin{figure*}
	\begin{center}
		\includegraphics[width=1\linewidth]{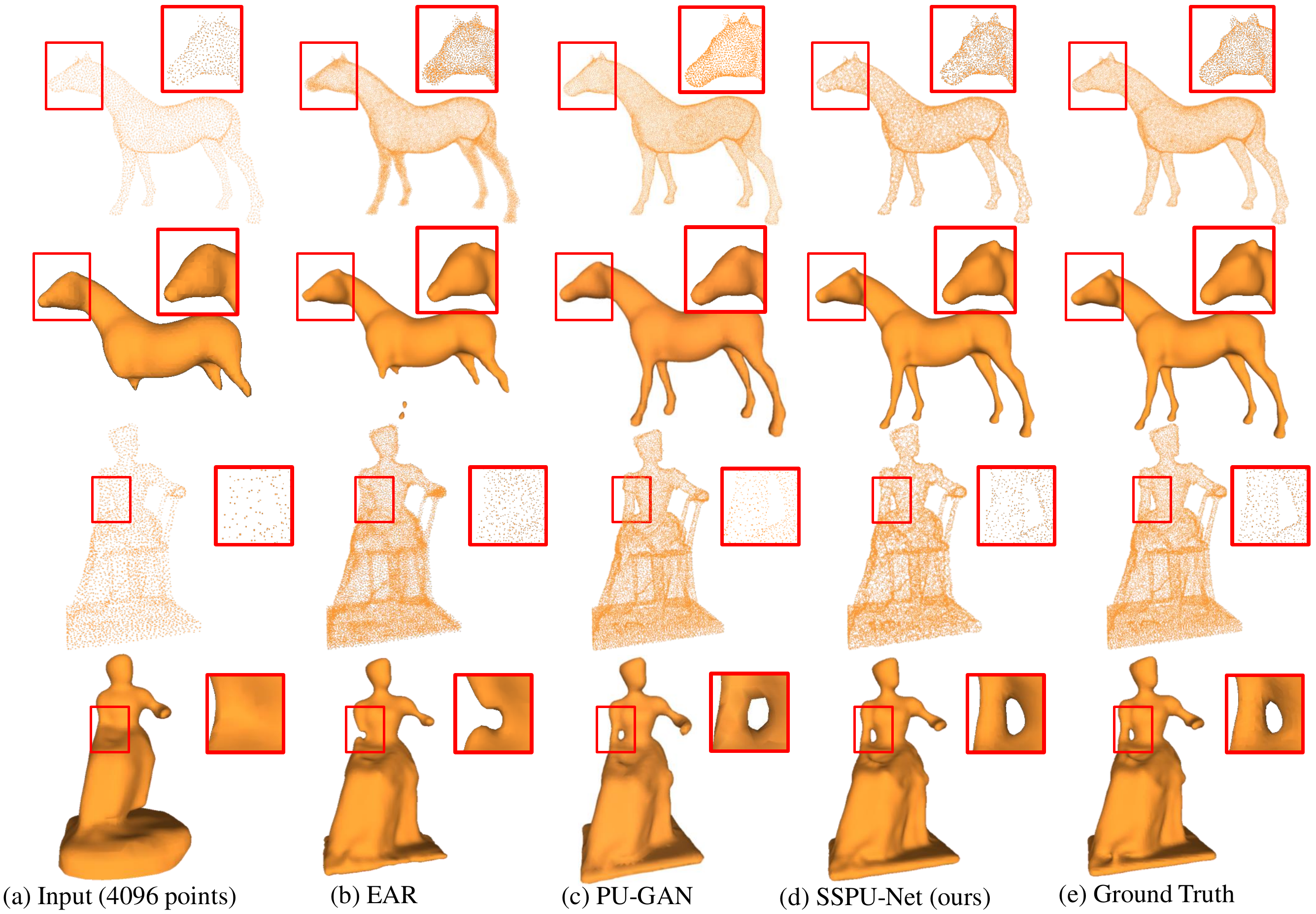}
	\end{center}
	\caption{Comparison results of point cloud upsampling ($\times$4) and surface reconstruction results with different methods (b-d) from the input point clouds (a).}
	\label{fig:comparing}
\end{figure*}

\begin{table}
	\centering
	\footnotesize
	\caption{The evaluation results ($\times4$) of different methods on the SSPU-DataSet. Note that CD, HD, and P2F are multiplied by 10$^3$. The best results are in bold, and second best results are underlined. ``\#'', ``+'' and ``*'' indicate that these methods are optimization-based, supervised and self-supervised methods, respectively.}
	\begin{tabular}{lcccc}
		\toprule
		Methods & \;\;\;CD\;\;\; & \;\;\;HD\;\;\; & P2F-mean & P2F-std\\
		\midrule
		\multicolumn{5}{l}{Input 2048 Points}\\
		EAR~\cite{huang2013edge}$^\#$ & 0.44  & 4.43 & 7.30 & 10.25\\
		PU-Net~\cite{pu-net}$^+$ & 0.34  & 3.84 & 7.27  & 8.05\\
		MPU~\cite{pppu}$^+$ & {\bf 0.23}  & {\bf 1.74} & {\bf 3.64} & {\bf 6.98}\\
		PU-GAN~\cite{pu-gan}$^+$ & \uline{0.24}  & 6.85 & \uline{3.74}  & 8.25\\
		SSPU-Net (ours)$^*$ & 0.29  & \uline{3.10} & 4.29 & \uline{7.29}\\
		\midrule
		\multicolumn{5}{l}{Input 4096 Points}\\
		EAR~\cite{huang2013edge}$^\#$       & 0.38  & 4.28 & 7.14 & 9.17\\
		PU-Net~\cite{pu-net}$^+$       & 0.24  & 3.38 &  7.36 & 8.06\\
		MPU~\cite{pppu}$^+$       & 0.18  & {\bf 1.34} & {\bf 3.23}  & {\bf 7.01}\\
		PU-GAN~\cite{pu-gan}$^+$       & {\bf 0.12}  & 9.53 & \uline{3.26}  & 7.48\\
		SSPU-Net (ours)$^*$ & \uline{0.16}  & \uline{1.83} & 3.72 & \uline{7.21}\\
		
		\bottomrule
	\end{tabular}
	\label{tab:comparisons2048}
\end{table}

\begin{table}
	\centering
	\footnotesize
	\caption{The evaluation results ($\times4$) of different methods on the PU-GAN~\cite{pu-gan} dataset. Note that CD, HD, and P2F are multiplied by 10$^3$.}
	\begin{tabular}{lccc}
		\toprule
		Methods & \;\;\;CD\;\;\; & \;\;\;HD\;\;\; & P2F-mean \\
		\midrule
		EAR~\cite{huang2013edge}$^\#$       & 0.52  & 7.37 & 5.82 \\
		PU-Net~\cite{pu-net}$^+$       & 0.72  & 8.94 &  6.84 \\
		MPU~\cite{pppu}$^+$       & 0.49  & 6.11 &  \uline{3.96} \\
		PU-GAN~\cite{pu-gan}$^+$       & {\bf 0.28}  & \uline{4.64} &  {\bf 2.33} \\
		SSPU-Net (ours)$^*$ & \uline{0.37}  & {\bf 3.47} & 4.43 \\
		\bottomrule
	\end{tabular}
	\label{tab:comparisons2048_onpugan}
\end{table}

We also conduct experiments on the dataset of PU-GAN~\cite{pu-gan} to demonstrate the effectiveness of our method. We show the evaluation results in Table~\ref{tab:comparisons2048_onpugan}. It can be found that the performance of our SSPU-Net is still superior to PU-Net, while the performance is comparable to PU-GAN and MPU. Note that the PU-GAN dataset is smaller than the collected SSPU-Dataset. Generally, self-supervised methods require a larger dataset to obtain better performance. The performance on those two datasets further demonstrates that our method can achieve good upsampling results in a self-supervised manner.

\textbf{Visual results.} In addition to the quantitative results, in Figure~\ref{fig:comparing}, we also provide the visualization results of the upsampling results and the surface reconstruction results. The first column (a) shows the input sparse point cloud (4069 points) and the corresponding surface reconstruction result, and the last column (e) shows the corresponding ground truth (16384 points). The columns (b), (c), (d) show the results of EAR, PU-GAN, and SSPU-Net (ours), respectively. It can be seen that our SSPU-Net can recover more details than other methods. For example, our method can more accurately recover the legs of the horse and arms of the statue. On the one hand, our method considers  the local geometric structures of the point cloud through two NEUs. As claim in~\cite{huang2013edge}, EAR is limited by the parameter radius, which measures the size of the neighborhood. As PU-GAN does not consider the local geometric structures of the point cloud, it cannot accurately recover the detailed local structures. On the other hand, although our method is self-supervised, the shape-consistent loss and image-consistent loss provide sufficient supervision signals to train the network. In addition, the good upsampling results also lead to the accurate surface reconstruction results. The visualization and reconstruction results further demonstrate the effectiveness of our method.

\begin{figure*}
	\begin{center}
		\includegraphics[width=1\linewidth]{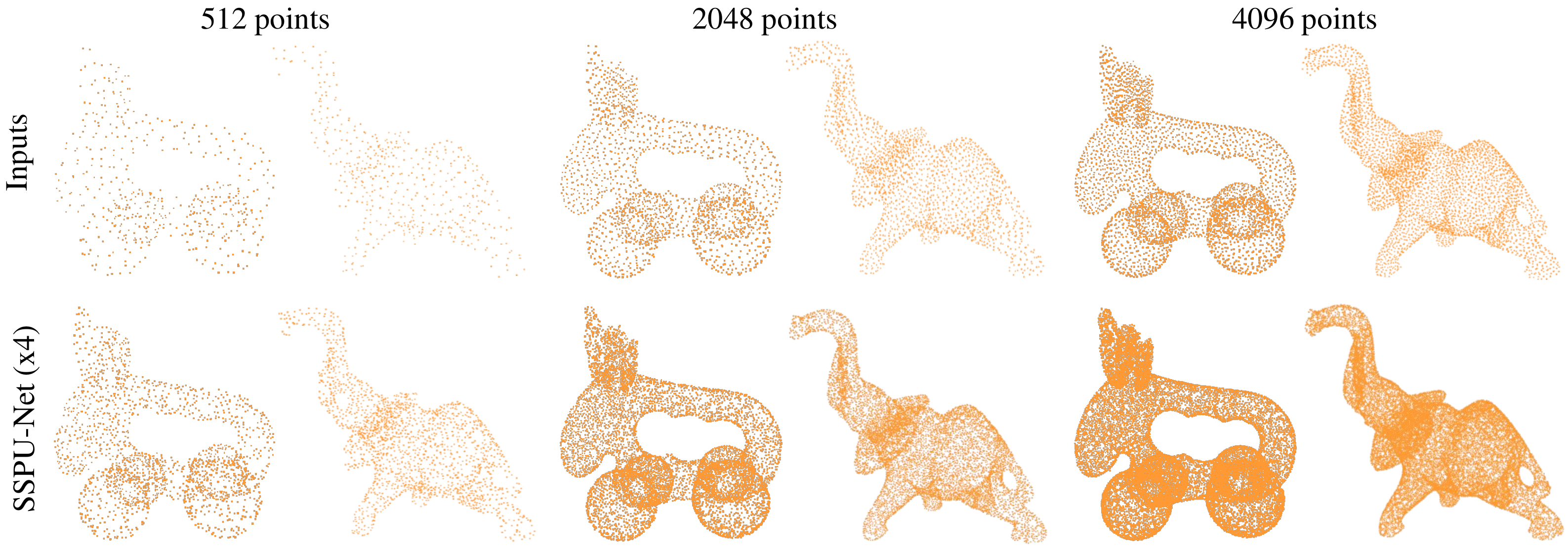}
	\end{center}
	\caption{The top and bottom rows show different inputs and the corresponding upsampling ($\times$4) results.}
	\label{fig:multi}
\end{figure*}


\begin{table}[htbp]
	\centering
	\footnotesize
	
	\caption{The evaluation results ($\times4$) of different components of SSPU-Net on the SSPU-Dataset. Note that CD, HD, and P2F are multiplied by 10$^3$.}
	\begin{tabular}{lcccc}
		\toprule
		&  CD & HD & P2F-mean & P2F-std\\
		\midrule
		w/o NEU      & \uline{0.35}  & \uline{4.38} & \uline{6.63} & \uline{7.70}\\
		w/o shape-consistent loss		& 2.31	& 27.74 &  19.73 & 14.85\\
		w/o image-consistent loss       & 1.11  & 23.98 &  14.83 & 13.95\\
		full pipeline & {\bf 0.29}  & {\bf 3.10} & {\bf 4.29} & {\bf 7.29}\\
		\bottomrule
	\end{tabular}
	\label{tab:ablation study}
\end{table}

\subsection{Ablation Study}


\textbf{Neighbor expansion unit.} To study the effect of the proposed NUE, we replace it with the corresponding structure (up-down-up expansion unit) in PU-GAN~\cite{pu-gan} to evaluate the performance on the SSPU-dataset. We report the results in Table~\ref{tab:ablation study}. Note that ``w/o NEU'' in Table~\ref{tab:ablation study} denotes the corresponding results. Although the improvement of the NEU module is not so large in comparison to the consistency loss, the improvement of the NEU is remarkable. For example, with the NEU, the upsampling performance is improved by 1\% and 2\% in terms of the HD and P2F-mean metrics. The NEU exploits the local geometric structure of the sparse point cloud to adaptively interpolate neighboring points so that the local structures of point clouds can be preserved in the upsampling process. The up-down-up expansion unit uses feature duplication to interpolate points, so it cannot capture the local geometric structures of the point cloud.

\textbf{Shape/Image-consistent loss.} As our method is self-supervised, the shape-consistent loss and the image-consistent loss are regarded as the supervision signals to train our network. In Table~\ref{tab:ablation study}, we report the evaluation results of discarding the shape-consistent loss or the image-consistent loss. As can be seen from the table, without the shape-consistent loss or the image-consistent loss, the performance will decrease significantly. The shape-consistent loss encourages the sparse and dense point clouds to have similar 3D shapes overall, while the image-consistent loss encourages the sparse and dense point cloud to have similar local geometric structures through multi-view images.

\textbf{Different number of input points.} To verify the effectiveness of SSPU-Net on different numbers of input points, we show the visualization results in Figure~\ref{fig:multi}. Note that we only train the network on 2048 points, and directly test the trained model on different numbers of input points, including 512, 2048, and 4096 points. It can be seen that our method can produce good upsampling results in the cases of different numbers of input points. The experiments show that our method is robust to different densities of input point clouds.


\begin{figure*}[htbp]
	\begin{center}
		\includegraphics[width=1\linewidth]{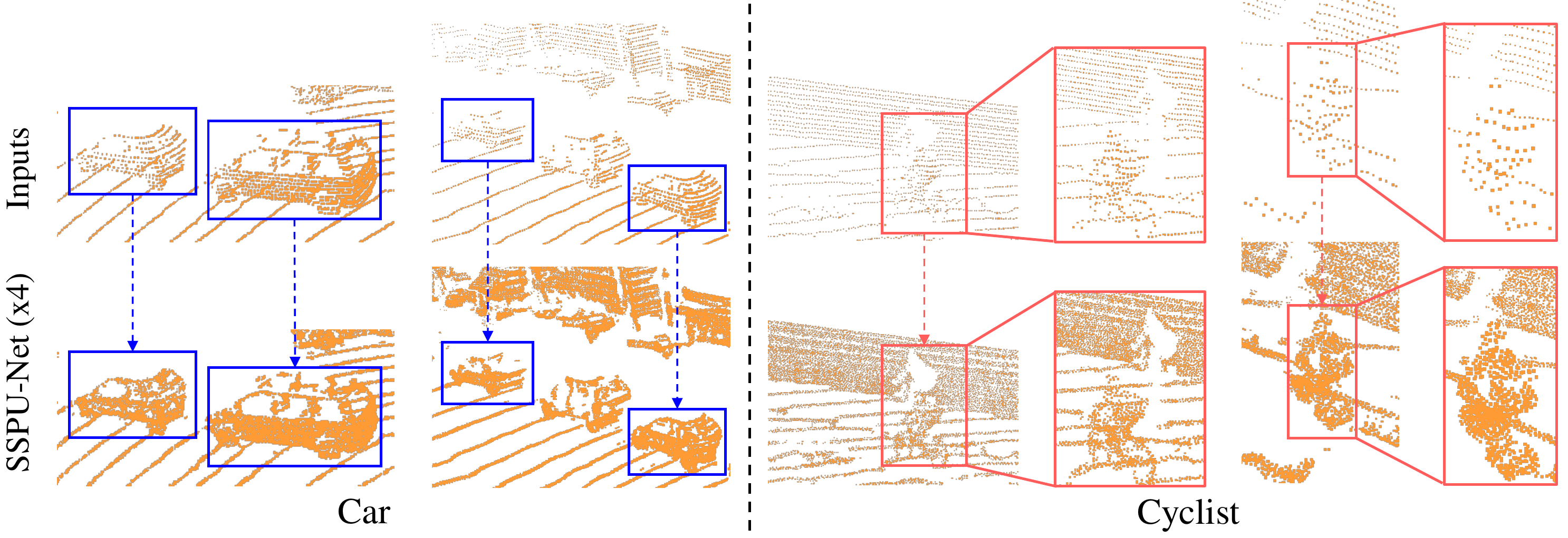}
	\end{center}
	\caption{The upsampling results of the point clouds of cars and cyclists in the KITTI dataset.}
	\label{fig:car_bike}
\end{figure*}

\textbf{Generalization ability.} Furthermore, we also directly test the pre-trained model (trained on the SSPU-Dataset) on the outdoor KITTI dataset~\cite{geiger2013vision} to demonstrate the generalization ability of the proposed method. As shown in Figure~\ref{fig:car_bike}, we visualize the upsampling results of the car and cyclist categories on the KITTI dataset. The upsampling results show that our SSPU-Net can still preserve the geometric details of the real scanned scenes. Thus, the upsampling results further demonstrate that our method has good generalization ability.

\subsection{Classification}
In this subsection, we perform the downstream classification task to demonstrate that the upsampling results can further improve the classification performance. Specifically, we use PointNet++~\cite{qi2017pointnet++} to conduct the experiments on the ModelNet40~\cite{wu20153d} dataset. We first use the pre-trained model to generate $\times$4 upsampling results (1024 points $\rightarrow$ 4096 points) on the ModelNet40 dataset. Then, we train the PointNet++ on the upsampled ModelNet40 dataset with the same experimental settings as the original PointNet++.

As shown in Table~\ref{tab:classification}, we report the classification results using different point cloud upsampling methods, including SSPU-Net, PU-Net, and PU-GAN. Compared with the original PointNet++, our SSPU-Net can improve the classification results with the gain of 1.6\% on the mean class accuracy. In addition, our method is superior to PU-Net and achieves good results, which are comparable to PU-GAN. The classification results demonstrate that our method can improve the performance of the downstream classification task.
\begin{table}[htbp]
	\centering
	\footnotesize
	\caption{The classification results (\%) on the ModelNet40 dataset. ``mAcc'' and ``OA'' denote the mean class accuracy and overall accuracy.}
	\begin{tabular}{lcc}
		\toprule
		Method &  \;\;\;mAcc\;\;\; & \;\;\;OA\;\;\; \\
		\midrule
		PointNet++      & 86.2  & 90.1\\
		PointNet++ ($\times$4 by PU-Net$^+$)     & 85.4  & 89.8\\
		PointNet++ ($\times$4 by PU-GAN$^+$)     & {\bf 87.9}  & \uline{90.2}\\
		PointNet++ ($\times$4 by SSPU-Net$^*$)       & \uline{87.8}  & {\bf 90.6}\\
		
		\bottomrule
	\end{tabular}
	\label{tab:classification}
\end{table}


\subsection{Retrieval}
Here, we also conduct the experiments on the point cloud based place recognition task~\cite{uy2018pointnetvlad,liu2019lpd,hui2021efficient} to demonstrate that our method can still improve the downstream retrieval task. In the experiment, we use the classical PointNetVLAD~\cite{uy2018pointnetvlad} to perform the experiments on the Oxford RobotCar dataset~\cite{maddern20171}. Specifically, we first use the pre-trained model to generate $\times$2 upsampling results (4096 points $\rightarrow$ 8192 points) of the Oxford RobotCar dataset. Then, we use the upsampled data to train the PointNetVLAD and evaluate the retrieval results.

In Table~\ref{tab:retrieval}, we list the retrieval results using different upsampling methods, including SSPU-Net, PU-Net, and PU-GAN. From this table, it can be seen that using upsampled data with SSPU-Net can effectively improve the performance of PointNetVLAD (dubbed as PN\_VLAD). Although our SSPU-Net is self-supervised, we can also yield comparable results to PU-GAN. In addition, since our SSPU-Net considers the local geometric structures of the point cloud, it can obtain better retrieval results than PU-Net. 






\begin{table}[htbp]
	\centering
	\footnotesize
	
	\caption{The retrieval results of the average recall (\%) at top 1\% (@1\%) and at top 1 (@1) on the Oxford RobotCar dataset.}
	\begin{tabular}{lcc}
		\toprule
		&  \shortstack{Ave recall @1\%} & \shortstack{Ave recall @1} \\
		\midrule
		PN\_VLAD     & 81.0  & 62.7 \\
		PN\_VLAD ($\times$2 by PU-Net$^+$) & 81.4  & 62.9 \\
		PN\_VLAD ($\times$2 by PU-GAN$^+$) & {\bf 82.1}  & \uline{63.4} \\
		PN\_VLAD ($\times$2 by SSPU-Net$^*$) & \uline{82.0}  & {\bf 63.5} \\
		\bottomrule
	\end{tabular}
	\label{tab:retrieval}
\end{table}


\section{Conclusion}
In this paper, we proposed SSPU-Net, an end-to-end self-supervised point cloud upsampling network. The proposed neighbor expansion unit (NEU) can adaptively learn weights from the local geometric structures of the point cloud for point interpolation. The proposed differentiable rendering unit (DRU) can render the input point cloud into multi-view images, and also support end-to-end training. In order to train the network in a self-supervised manner, we proposed a shape-consistent loss and an image-consistent loss. The shape-consistent loss encourages the input sparse point cloud and the generated dense point cloud to have similar 3D shapes, and the image-consistent loss encourages the rendered images of the input and generated point clouds to have similar local geometric structures. Experimental results on the point cloud upsampling task and downstream classification and retrieval tasks demonstrate the effectiveness of our proposed self-supervised point cloud upsampling network.


\begin{acks}
This work was supported by the National Science Fund of China under Grant 61876084.
\end{acks}

\bibliographystyle{ACM-Reference-Format}
\bibliography{acmart}

\end{document}